# Problems of Non-equivalent Words in Technical Translation

**Mohammad Ibrahim Qani**

Assistant Professor (Pohanyar), Faryab University, Afghanistan. 2022

**Abstract:** translating words which do not have equivalent in target language is not easy and finding proper equivalent of those words are very important to render correctly and understandably, the article defines some thoughts and ideas of scientists on the common problems of non-equivalent words from English to Russian language and includes English and Russian examples and ideas of certain scientist. The English language is worldwide spoken and there are 1.35 billion English speakers and over 258 million Russian speakers according to the 2021's statistics. Inevitably, these billions of speakers around the world have connection and they may have deal in different criteria. In order to understand one another they need to have a pure and fully-understood language. These pure languages understanding directly relates to translation knowledge where linguists and translators need to work and research to eradicate misunderstanding. Misunderstandings mostly appear in non-equivalent words because there are different local and internal words like food, garment, cultural and traditional words and others in every notion. Truly, most of these words do not have equivalent in the target language and these words need to be worked and find their equivalent in the target language to fully understand the both languages. However, some of these non-equivalent words are already professionally rendered to the target language but still there many other words to be rendered. Hence, this research paper includes different ways and rules of rendering non-equivalent words from source language to the target language.

**Keywords:** translation, English – Russian non-equivalence, non-equivalent vocabulary, alternative of non-equivalent vocabulary, challenges in translating.

**Aim of the Study**

The role of non-equivalence in technical translation and finding the appropriate problems and difficulties of the field, providing scientific guidance to solve the issue, finding the available challenges and presenting solution in a proper way, stating translators lacking and troubles in translating a non-equivalent word from the source language in the target language and some other challenges and solutions considering technical translation.

**Methods of Research**

In this article, I used the APA style and the usual and old research method which is called Library Research (data collection), online libraries, PDF books, available research papers in reliable journals and websites, and some other sources. This article covers the problems of non-equivalent words in technical translation and properly rendering from the source language into the target language.

**Introduction**

Translation plays crucial and important role in today's world and non-equivalence is one of the most important parts and elements of transition. The improvement of technology has brought close the people around the world and they are simply able to connect to one another. Most of the businesses are structured online and the business peculiarities are also relying on online services. We are witnesses of many online events nowadays. The universities, trade centers, governmental bureau, social and cultural development centers, youth associations, foundation, governmental and non-governmental organizations and other institutes are holding their conferences, deals, inaugurations, scientific gatherings some other events through online facilities of technology and we know that these events conduct in different languages. They need to understand every moment and words of such events. In order to understand all the discussion we need to have equivalent of every non-equivalent words into different languages. Therefore, I tried to compile important and useful theories, perspective and common problems in translating non-equivalent words from source language into target language, in this article you can find effective rules to escape from ambiguity. However, these rules and regulations are well-known among translators and interpreters around the world and they have the capabilities to regard the rules in understand listeners in interpreting or translating events but it is not enough to have a clear communication worldwide. It would be better if other people have at least the general information about non-equivalent rendering rules. Therefore, this article aims to explain and simplify major difficulties and challenges of non-equivalent words. The universal features of language material desires to describe world picture of different language speakers. Therefore, rendering non-equivalent words of different languages and finding their difficulties will help people around the world to have pure communication and easily understand each other.

## The Common Problems of Non-equivalent

The more common types of non-equivalent problems and difficulties for the translators and some attested strategies for dealing with them is divided in some factors: a) a word of warning b) extra linguistics. The problem of non-equivalent has been drawing the attention of many researchers. Jakobson claims that "there is ordinarily no full equivalence between code units" Jakobson also explains the differences between structures, terminology, grammar and lexical forms of languages are the main reasons of non-equivalence. Jacobson states that "equivalence in difference is the cardinal problem of language and the pivotal concern of linguistics." In his theory, the general principle of cross-language difference and the concept 'semantic field' has been established (Jakobson, 1959: P 252).

Catford found that there are two factors which affected the equivalence i.e. linguistic and cultural factors, leading to two kinds of equivalents i.e. linguistic and cultural equivalents. This finding of Catford is very significant because it consists of both important approaches toward equivalence, namely, linguistic and cultural approaches. On the contrary, there were some arguments against Catford theory. Snell-Hornby claims that textual equivalence introduced by Catford is "circular" and his examples are "isolated and even absurdly simplistic". Furthermore, she criticizes equivalence in translation is an illusion because there are many aspects, including textual, cultural and situational ones, get involved in the equivalent degree of the translation. House also agrees that not only functional but situation factor need to be taken into consideration during the process of translation (Catford, 1965: PP. 8, 191).

Equivalent effect, as judged by Newmark, is "the desirable result, rather than the aim of any translation". Accordingly, the equivalent effect is a result which all translators long to achieve. Further, Newmark argues that the text may reach a 'broad equivalent effect' only if it is 'universal' that means cross culture share common ideas (Newmark, 1988: pp. 5, 134).

As Researchers indicate (I.M.Vereshagin, V.G. Kostomarov and others), that in order to establish availability of national cultural specifics of meaning of a word can be done through comparison the semantics of words of two languages (or more). Comparative Researches of languages showed that national cultural differences appear especially on the lexical phraseological levels.

A number of researchers agree that in Translation studying of non-equivalent vocabulary is linked with the notion of "transferability" and "equivalent", with the problem of non-equivalent and vocabulary translation means which denotes items or phenomena of national culture.

Classification of non-equivalent vocabulary can be conducted by genetic trait.
1. Word of life (all neologisms)
2. Names of items and phenomena's of traditional life.
3. Historicisms
4. Lexis of phrasedogical units
5. Folklore words
6. Slang words/youth slang, criminal slang, military slang, any professional slang
7. Social-political vocabulary
8. Reduced, colloquial vocabulary.

A.O. Ivanov divides all non-equivalent vocabulary into three big groups.
1. Referentially-non-equivalent, which includes term, individual (author), neologisms, semantic lacunas, words of wide semantics, complex words;
2. Pragmatically-non-equivalent, uniting abnormalities, foreign inclusions, abbreviations, words with suffixes of subjective evolution, interjections, imitation a sound and associative lacunas;
3. Alternatively-non-equivalent vocabulary including proper names, circulation, realia and phraseologisms.

Non-equivalence happens at word level. It means that target language (TL) has no direct equivalence for a word which occurs in the source language. There are many possible problems of non-equivalence between two languages. Non-equivalence occurs when the message in the source language is not transferred equally to the target language (Ivanov, 2007: pp. 1, 117).

**The problem of non-equivalence at word level of technical translation**

Baker states that non-equivalence at world level in technical translation means that the target language has not direct equivalent for a word which occurs in the source text. The lack of equivalence at word level poses the translation problems arising. She further unpacks this statement by asking a question; what does a translator do when there is no word in the target language which expresses the same meaning as the source language word?

The type and level of difficulty posed can vary tremendously depending on the nature of non-equivalence. Different kinds of non-equivalence require different strategies, some very straightforward, others more involved and difficult to handle.

These are some common problems of non-equivalence at word level of technical translation: culture-specific concepts, the source-language concept is not lexicalized in the target language; the source-language is semantically complex; the source and target languages make different distinction in meaning; the language lacks a superordinate; the target language lacks a specific term (hyponym); difference in physical or interpersonal perspective; differences in expressive meaning; differences in form; differences in frequency and purpose of using specific forms; the use of loan words in the source text.

Translation whether defined as a study and process about re-express the message and meaning of source language into the appropriate equivalent of target language. The concept of 'equivalence' is introduced in the definition. It means that the target language has direct equivalent for a source language word. However, there are many occasions in which non-equivalence at word level in technical transition accuse between the two languages.

Strategies used for dealing with non-equivalence at word level in technical translation are translation by more general word (superordinate); translating by more neutral/less expressive word; translating by cultural substitution; translating using a loan word or loan word plus explanation; translating by paraphrase using a related word; translating by paraphrase using unrelated word; translating by omission; translating by illustration.

According to Mona Baker, non-equivalence at word level means that the target language has no direct equivalent for a word which occurs in the source text. (Baker, 1992: pp. 9, 3-4)

**Culture-specific concepts:** Based on this problem, the source-language word may express a concept that is totally unknown in the target language culture. The concept may be abstract or concrete; it may relate to a religious belief, a social custom, or even a type of food. For example, the word **privacy** is a very "English" concept, which is rarely understood by people from other cultures. The source language word may express a concept which is totally unknown in target language. The concept in question may be abstract or concrete; it may relate to a religious belief, a social custom, or even a type of food. Such concepts are often referred to as' culture-specific. The source-language concept is not lexicalized in the target language. The source language word may express a concept which is known in the target culture but simply not lexicalized, that is not 'allocated' a target language word to express it. For example, in Russian the word 'Дача' has not direct equivalent in English although it can be understood as a house in a village and nature.

It is possible to come across a word which communicates a concept in the source target that is unknown in the target culture. This concept could be abstract or concrete; it could refer to a social custom, a religious belief, or even a type of food. The source language may be semantically complex. This is a fairly common problem in translation. Words do not have to be morphologically complex to be semantically complex. In other words, a single word which consists of a single morpheme can be sometimes expressing a more complex set of meanings than a whole sentence. Languages automatically develop very concise forms for referring to complex concept if the concepts become important enough to be talked about often. (Zokirova, 2016: pp. 88-89)

**The Source-Language Concept is not Lexicalized in the Target Language:** This problem occurs when the source language expresses a word which easily understood by people from other culture but it is not lexicalized. For example, the word **savoury** has no equivalent in many languages, although it expresses a concept which is easy to understand. It means that a concept that is known by people in some areas does not always have the lexis in every area. (Larson, 1984: p. 145)

**The Source Language Word is Semantically Complex**: The source-language word can be semantically complex. This was fairly common problem in translation. Words did not have to be morphologically complex to be semantically complex. In other words, a single word which consisted of a single morpheme could sometimes express a more complex set of meaning than a whole sentence.

**The Source and Target Languages Make Different Distinctions in Meaning**: What one language regards as an important distinction in meaning another language may not perceive as relevant. The target language may make more or fewer different distinction in meaning than the source language (Widhiya, 2010: pp. 3, 181-182).

**Differences in expressive meaning:** There may be a target language word which has the same proportional meaning as the source word, but it may have different expressive meaning the difference may be considerable or it may be subtle but important enough to pose a translation problem in a given context. It is usually easier to add expressive meaning then to subtract it. In other words, if the target language equivalent is neutral compared to the source language item, the translator can sometimes add the evaluative elements by means of a modifier or adverb if necessary, or by building it in somewhere else in the next. Differences in expressive meaning are usually difficult to handle when the target language item. This is often the case with items which relate to sensitive issues such as region, political, and sex.

**Differences in forms:** There is often no equivalent in the target language for a particular form in source text. Certain suffixes and prefixes which convey propositional and other types of meaning in Russian often have no direct equivalents in English. It is most important for translator to understand the contribution that affixes make to the meaning of words and expressions, especially since such affixes are often use creatively in English to coin new words for various reasons, such as filling temporary sematic gaps in the language and creating humor, their contribution is also important on the area terminology and standardization (Pham, 2010: PP. 10, 112).

**The use of loan words in the source text:** The use of loan words in the source text poses a special problem in translation. Different in the original meaning, loan words usually are used to show prestige. This matter is often impeding in term of translation, which is caused by there are no equivalent words in target language. (House, 2002: P. 15)

**The Use of Loan Words in the Source Text:** Once a word is loaned into a particular language, we cannot control its development or its additional meaning. For example, **dilettante** is a loan word in English, Russian, and Japanese; but Arabic has no equivalent loan word. This means that only the prepositional meaning of dilettante can be rendered into Arabic; its stylistic effect would almost certainly have to be scarified. Loan words also pose another problem for the unwary translator namely the problem of false friends, or **faux amis** as they are often called in mentioned way. Translators should be more careful when they face the loan words in the process of translating a text.

In the English-speaking translation studies there are two more translation strategies not having the equivalents in the Russian language. They are **direct translation and oblique translation** which have been introduced into the science by the linguists J.-P. Vinay и J. Darbelnet. The classification offered by the French translation studies theorists can be considered to be quite a detailed one. These two strategies include seven methods, namely, the notion **direct translation** includes:

- ➢ borrowing (заимствование)
- ➢ calque (калькирование)
- ➢ literal translation (дословный, буквальный перевод).

Oblique translation consists of:
- transposition (транспозиция)
- modulation (модуляция)
- equivalence (эквивалентность)
- adaptation (адаптация)

For the term **direct translation,** the authors suggest to use its synonym (**literary**). As for the term **oblique translation** according to the authors' opinion it does not have the equal meaning with **free translation**.

As for the presence of the term **oblique translation** in the Russian term system of translation studies, here we come across the lexical gap, for example, the only meaning which can be found for the word combination oblique translation in the Russian language relates to the field of mathematics and the meaning («наклонный перенос») has nothing to do with the theory of translation. In the English-Russian dictionaries the terminological word combination has not been included yet.

1. If a specific linguistic unit in one language carries the same intended meaning/message encoded in a specific linguistic medium in another, then these two units are considered to be equivalent. Equivalence is considering the essence of the translation.
2. The choice of a suitable equivalent in a given context depends on a wide variety of factors. Some of these factors may be strictly lingusitcs (ex. collocations and idioms), others may be extra linguistics (ex. pragmatics). Non-equivalence means that the target language has no direct equivalence for a word which occurs in the source text. The type and level of difficulty posed can vary tremendously depending on the nature of non-equivalence.
3. The particular issues addressed are the bi- directionality of translation equivalence, the coverage of multi-word units, and the amount of implicit knowledge presupposed on the part of the user in interpreting the data. Non-equivalence is a fact among languages.
4. Translators constantly face countless cases of more straightforward and clearer examples of non-equivalence in translation. When this happens they manage to translate and not only to "transfer" the words. An adequate approach to deal with cases of non-equivalence would be to use a combination of translation strategies (Yvonne Malindi, 2015: PP. 10, 55).

**Conclusion**

Translation is becoming crucial element and part of our life. Today's world extremely need arts of translation because the world today is getting as close as possible; people around the world have been trying to conduct better connection and relation among each other. Technology unexpectedly has developed and connected people around the world. Therefore, translation is most important tool in order to solve communicative challenges and difficulties among people around the world. Hence, non-equivalence is one of the most important parts of transition to escape from ambiguity and it helps to understand opponent clearly. Many updated sciences are available into different languages and the only tool that can help us to benefit from that information is translation. On the other hand, powerful societies and governments are working to influence other communities. They are donating fully-funded scholarships and other plans to develop their language, tradition, culture and other social activities. These activities can be done by the help of translation and the term of non-equivalent words are one of the essential parts of translation where it can help the language learners to understand the scope properly and authentically. I have just revolved about available problems and difficulties of non-equivalent words. There are scientific methods to render those words properly and understandably from source language to the target language. It is really necessary to learn those who are working as a translator of interpreter because without those methods it is too hard to render a word have no equivalent to the target language. If we were not able to translate those words properly we translation process can be meaningless. Understanding knowledge of non-equivalent words in technical translation is extremely necessary and these scientific methods of rendering words from source language to the target language help the translator, transcriptionist, editor, proofreader, interpreter and others to convey their words clearly and understandably from their native language to the target language.

**References**

1. Ivanov A.O. (2007). Equivalent Vocabulary. Voronezh, Russian, Spb.: Soyuz
2. Zokirova S.M. (2016). The notion of non-equivalence Vocabulary in Translation, IJSELL.
3. Jakobson, Roman (1959). 'On Linguistic Aspects of Translation', in R. A. Brower (ed.) *On Translation,* Cambridge, MA: Harvard University Press, p. 232.

4. Baker, M. (1992). In Other Words: A Course Book on Translation. London: Routledge.
5. Catford, J. C. (1965). A Linguistic theory of translation; An Essay In Applied.
6. Newmark, P. (1988). A Textbook of Translation, U.K. Prentice Hall International Ltd.
7. Widhiya, Ninsiana. (2010). Problem Solving of Non-equivalence Problems in English, IAIN.
8. Pham, Thanh Binh. (2010). Strategies to Deal with Non-equivalence at Word Level in Translation, Hanoi University.
9. House, J. (2002). Universality versus culture specificity in translation 'Alessandra Riccardi' Translation studies: perspectives on an emerging discipline. UK.
10. Yvonne Malindi, Nokuthula. (2015). Solving Non-equivalence Problems when Translating Technical Text. University of Pretoria. Pretoria, Africa.
11. Larson, L. Mildred. (1984). Meaning Based Translation; A Guide to Cross Language Equivalence. Lanham: University Press of America.